\documentclass[10pt,twocolumn,letterpaper]{article}

\usepackage{cvpr}              

\makeatletter
\DeclareRobustCommand\onedot{\futurelet\@let@token\@onedot}
\def\@onedot{\ifx\@let@token.\else.\null\fi\xspace}
\def\eg{\emph{e.g}\onedot} 
\def\ie{\emph{i.e}\onedot}

\newcommand{\Eref}[1]{Eq.~\textcolor{blue}{\ref{#1}}}

\usepackage[dvipsnames]{xcolor}
\definecolor{grey}{rgb}{0.52,0.52,0.52}

\usepackage{comment}

\usepackage{times}
\usepackage{epsfig}
\usepackage{graphicx}
\usepackage{amsmath}
\usepackage{amssymb}
\usepackage{verbatim}
\usepackage{hhline}
\usepackage{multirow}
\usepackage{makecell}

\usepackage{epsfig}
\usepackage{graphicx}
\usepackage{amsmath}
\usepackage{amssymb}

\definecolor{cvprblue}{rgb}{0.21,0.49,0.74}
\usepackage[pagebackref,breaklinks,colorlinks,citecolor=cvprblue]{hyperref}

\def\paperID{218} 
\def\confName{3DV\xspace}
\def\confYear{2024\xspace}

\title{Stable Surface Regularization for Fast Few-Shot NeRF}

\author{Byeongin Joung$^1$, Byeong-Uk Lee$^2$, Jaesung Choe$^1$, Ukcheol Shin$^3$, Minjun Kang$^1$, Taeyeop Lee$^1$\\ In So Kweon$^1$, Kuk-Jin Yoon$^1$ \\
\\
KAIST$^1$ \quad KRAFTON$^2$ \quad CMU$^3$
}

\begin{document}
\maketitle
\begin{abstract}
This paper proposes an algorithm for synthesizing novel views under few-shot setup. The main concept is to develop a stable surface regularization technique called Annealing Signed Distance Function (ASDF), which anneals the surface in a coarse-to-fine manner to accelerate convergence speed. We observe that the Eikonal loss -- which is a widely known geometric regularization -- requires dense training signal to shape different level-sets of SDF, leading to low-fidelity results under few-shot training. In contrast, the proposed surface regularization successfully reconstructs scenes and produce high-fidelity geometry with stable training. Our method is further accelerated by utilizing grid representation and monocular geometric priors. Finally, the proposed approach is up to 45 times faster than existing few-shot novel view synthesis methods, and it produces comparable results in the ScanNet dataset and NeRF-Real dataset. 
\end{abstract}    
\section{Introduction}
\label{sec:intro}

Recently, in a way to encode implicit scene appearance and geometry, Neural Radiance Fields (NeRF)~\cite{mildenhall2021nerf} has been introduced.
It utilizes Multi Layer Perceptrons (MLPs) and bulk of training images to obtain radiance and volume information for each 3D coordinates.

NeRF has shown its effectiveness and potential in retrieving scene geometry, and was applied to various fields such as novel view synthesis~\cite{mildenhall2021nerf}, surface reconstruction~\cite{wang2022go}, dynamic scene rendering~\cite{pumarola2021d}, and lighting~\cite{rudnev2022nerf}.
However, since NeRF training depends on the amount of images from various viewpoints, building a neural implicit field with fewer input images is challenging.
Moreover, its nature of scene-by-scene optimization may take over 10 hours, making it difficult to apply in real-world applications.

\begin{figure}
    \centering
    \includegraphics[width=\linewidth]{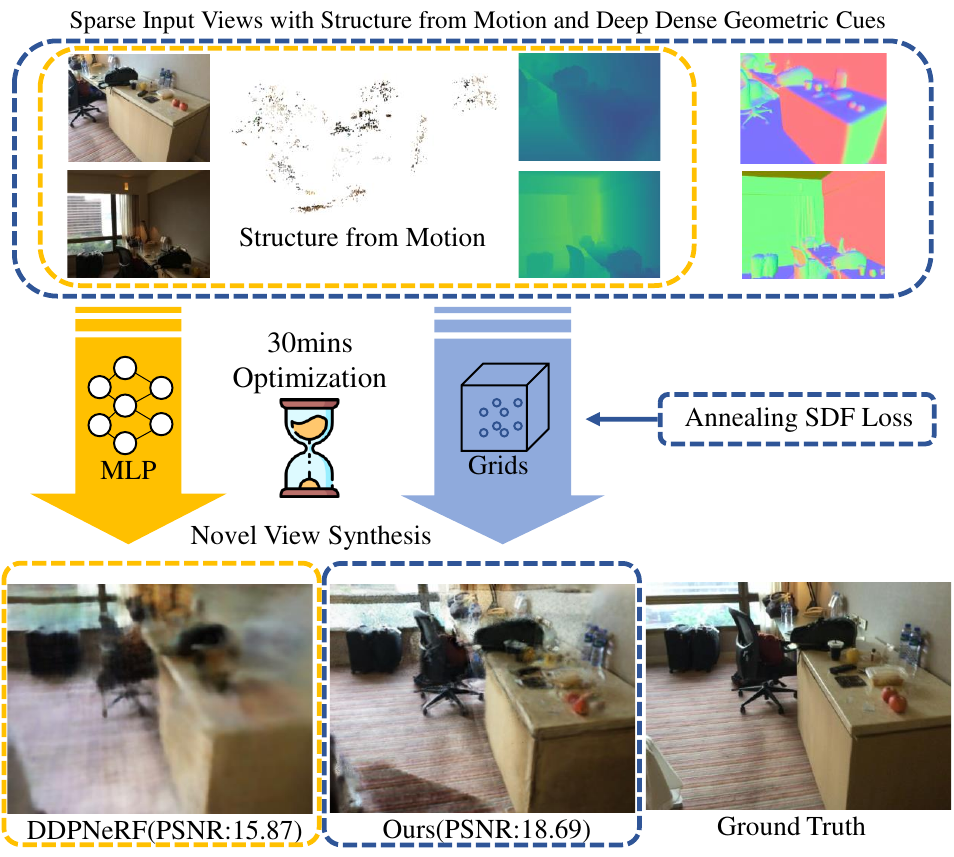}
    \caption{Our method can synthesize novel views within 30 minutes by utilizing multi-level voxel grid optimization. To overcome the limitation of novel view synthesis with sparse input images, we utilized additional strong geometric cues and our novel geometric smoothing loss, \textit{Annealing SDF loss.}}
    \label{Fig:main}
    \vspace{-3mm}
\end{figure}

\begin{figure*}[ht]
    \centering
    \includegraphics[width=\textwidth]{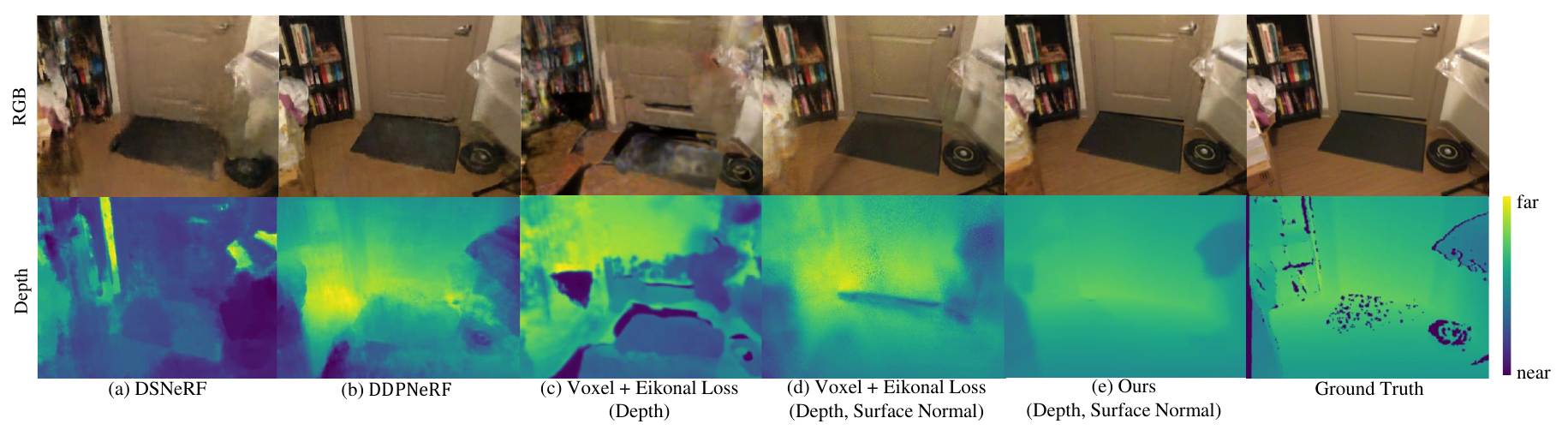}
    \caption{
        Novel view synthesis comparisons with state-of-the-art works~(a) DSNeRF~\cite{deng2022depth} (b) DDPNeRF~\cite{roessle2022dense} (c) voxel grids based approach~\cite{sun2022direct} with eikonal loss~\cite{gropp2020implicit} and depth supervision, and (d) ours. Top row: color images, bottom row: depth maps.
    }
    \label{Fig:intro_ver0}
    \vspace{-3mm}
\end{figure*}

To overcome issues in few-shot setups, series of studies utilized unobserved viewpoint regularization~\cite{niemeyer2022regnerf}, entropy minimization~\cite{kim2022infonerf}, and geometric priors~\cite{wei2021nerfingmvs,deng2022depth, roessle2022dense}.
%
%
Along with the strategies to handle issues in few-shot setups, voxel representation~\cite{sun2022direct, fridovich2022plenoxels} and hash encoding~\cite{muller2022instant} present the remarkable fast training for novel-view synthesis task. However, former approaches require more than 10 hours for training, while latter still struggle with sparse input views.

Up to our knowledge, MonoSDF~\cite{yu2022monosdf} is the first approach to provide analysis on fast few-shot NeRF training.
It studied MLPs and voxel grid optimization using off-the-shelf geometric cues with the Eikonal loss~\cite{gropp2020implicit}, which is known to be effective in Signed Distance Function (SDF)-based training. However, their study revealed that training fast few-shot NeRF with voxel grid optimization may struggle due to geometrically ambiguous regions driven by sparse input views.

Similarly, we observe that the naive application of the implicit geometric regularization, such as the Eikonal loss, fail to extract reliable color and geometry information in few-shot NeRF setting as visualized in Figure~\ref{Fig:intro_ver0}-(a, b, c). Especially, Figure~\ref{Fig:intro_ver0}-(c) shows that utilizing Eikonal loss in a voxel-based few-shot environment is not beneficial, as it is difficult to capture geometric information with sparse inputs and tends to converge to incorrect local minima. To address this issue, we propose a novel method for stable surface regularization in fast few-shot setup, called the Annealing Signed Distance Function~(ASDF) loss. This loss enforces strong geometric smoothing in the early stage of training, thereby enables coarse-to-fine surface regularization and promote stable convergence. By incorporating the ASDF loss, our approach shows stable optimization even in challenging conditions for learning geometry.

By jointly incorporating dense 3D predictions, multi-view consistency and our ASDF loss, our proposed algorithm successfully synthesize a novel view from few-shot training, while maintaining fast training time. We show brief overview in Figure~\ref{Fig:main}.

Our contributions can be summarized as follows:
\begin{itemize}
\itemsep0em
    \item Tackle the problem of the Eikonal loss when only a limited number of training views are available.
    \item Introduce a novel method to regularize surfaces in fast and few-shot neural rendering.
    \item Demonstrates comparable performance with up to 30$\sim$45x faster training speed.
\end{itemize}

\begin{figure*}[ht]
    \centering
    \includegraphics[width=0.90\textwidth]{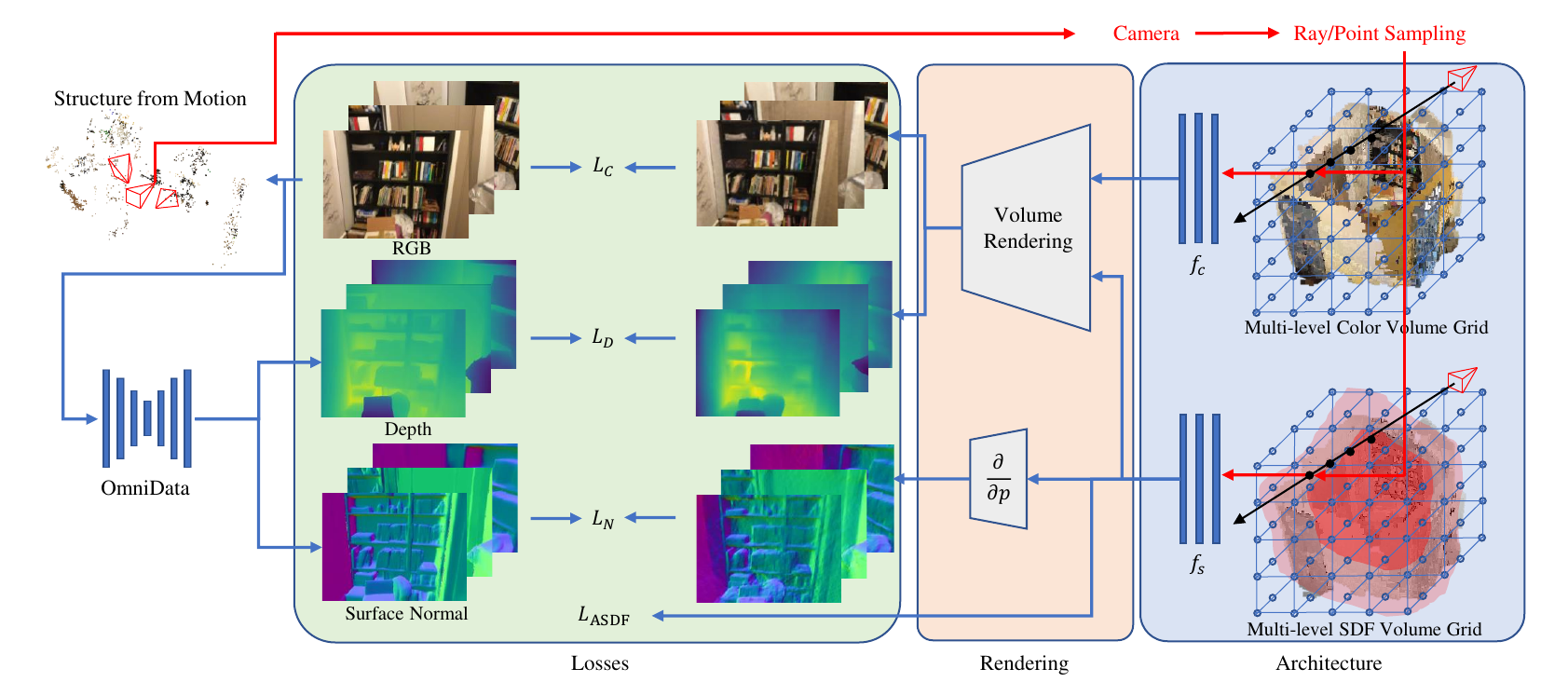}
    \caption{
        The overall pipeline. The proposed architecture utilizes structure from motion to extract sparse 3D information and camera poses from sparse input views, while off-the-shelf depth and surface normal are obtained from a pretrained network~\cite{eftekhar2021omnidata}. Points are sampled with respect to the camera from the structure from motion, and the feature value at corresponding points is extracted. SDF values, gradient of SDF values, and RGB values are decoded by simple MLP decoder, and RGB, surface normal, and depth values are extracted along the ray using volumetric rendering. The rendered RGB, depth map, and surface normal are supervised with their respective label using the loss functions $L_C, L_D$, and $L_N$, while SDF values are supervised with the loss functions $L_{\text{ASDF}}$ which is composed by $L_{\text{GS}}$, and $L_{\text{wEik}}$.        
    }
    \vspace{-2mm}
    \label{Fig:architecture}
\end{figure*}

\section{Related Works}

\noindent  \textbf{Neural Radiance Fields.} \
A newly introduced field, neural implicit representation~\cite{mildenhall2021nerf}, has rapidly developed in recent past. In order to enhance the capabilities of the neural implicit representation, several approaches have been proposed in recent years. For instance, \cite{wang2021ibrnet, chen2021mvsnerf, nguyen2022hrf} utilized multi-view consistency to improve generalization, while \cite{lin2021barf} proposed an optimization method based on a bundle of camera poses, which have errors. \cite{wang2021neus, yariv2021volume, zhang2021learning, oechsle2021unisurf, fu2022geo} show that the surface reconstruction can be adapted to neural implicit representation and help implicit surface representation learning. It is also adapted to 3D scene reconstruction with neural implicit representation~\cite{azinovic2022neural, wang2022go}. To reduce the training time of the optimization, \cite{sun2022direct, fridovich2022plenoxels} utilized voxel grids and \cite{muller2022instant} utilized hash encoding.

\noindent  \textbf{Few-shot NeRF.} \
Despite this success, NeRF requires dense inputs for high-quality novel view synthesis and shows limited performance with sparse inputs. One of the promising methods to overcome this issue is to use geometric constraints with NeRF for regularization. RegNeRF~\cite{niemeyer2022regnerf} introduced patch-based geometry and color regularization, InfoNeRF~\cite{kim2022infonerf} proposed ray entropy minimization and information gain reduction across different viewpoints. 
While previous studies, InfoNeRF and RegNeRF, augment the training rays by sampling unseen neighbor views, our ASDF is designed to learn surface geometry in coarse-to-fine manner.
SinNeRF~\cite{xu2022sinnerf} uses pseudo geometry and color in unseen view for regularization. More recently, DiffusioNeRF~\cite{wynn2023diffusionerf} uses a denoising diffusion model and learns to regularize color and depth within the patch.

\noindent  \textbf{NeRF with geometry prior.} \
On the other way, additional constraints are adapted to NeRF. \cite{liu2020neural, neff2021donerf, wei2021nerfingmvs, wang2022neuris} show geometric priors can be adapted to neural implicit representation. DONeRF~\cite{neff2021donerf} proposed a method that utilizes a monocular depth oracle network to sample points along the implicit surface to boost optimization time and enhance performance by reducing redundant sampling. NerfingMVS~\cite{wei2021nerfingmvs} also utilized adapted depth priors to sample points in indoor multi-view stereo (MVS) with structure from motion(SfM). NeuRIS~\cite{wang2022neuris} shows surface normal that can achieve stable optimization in large textureless regions. 
SCADE~\cite{uy2023scade} utilized monocular depth estimation while addressing its inherent ambiguities.
Besides on the other works, by utilizing full information by multi-view consistency and geometric cues from sparse input views with our strong geometric smoothing loss, we successfully address limitations of NeRF.

\section{Preliminary}

\noindent \textbf{NeRF}~\cite{mildenhall2021nerf} is a pioneering paper that encodes appearance and geometry of a scene within MLPs network to synthesize novel views. Given posed images, the MLPs are trained to emit color~$\mathbf{c}$ and density~$\sigma$ at given query point~$\mathbf{p}$ as:
\begin{equation}\label{eq:nerf}
    [\mathbf{c}; \sigma] = f(\mathbf{p}, \mathbf{d}),
\end{equation}
where $\mathbf{d}$ is the viewing direction and $f(\cdot)$ indicates the MLPs network. Then, this paper propose differential volumetric rendering to synthesize color value as:

\begin{equation}\label{eq:volumetric rendering using density}
    \hat{C}_r = \sum_{i=1}^{N} T_i\alpha_i\mathbf{c}_i~~\text{s.t.}~~\alpha_i=1{-}\exp({-}\sigma_i\delta_i),
\end{equation}
where $\hat{C}_r$ is the rendered color at the ray~$r$, $T_i$ is transmittance defined by $T_i = \text{exp}\left( -\Sigma _{j=1}^{i-1}\sigma_j \delta_j \right)$ and $\delta_i$ = $t_{i+1} - t_i$ is the distance between sampled adjacent query points.

\noindent \textbf{NeuS}~\cite{wang2021neus} proposes a way of 3D reconstruction from differential volumetric rendering as in~\Eref{eq:volumetric rendering using density}. Instead of using density~$\sigma$, this paper designs a unique way of learning SDF as:

\begin{equation}
\alpha(\mathbf{p}) = \max\left( \frac{-\frac{d\Phi_s}{dt}(s(\mathbf{p}))}{\Phi_s(s(\mathbf{p}))}, 0\right),
\end{equation}
where $\alpha(\mathbf{p})$ is opaqueness at point~$\mathbf{p}$. Here, $\Phi_s$ is a sigmoid function and $s(\mathbf{p})$ is the SDF value for the point $\mathbf{p}$. The surface normal value can be predicted by differentiating the SDF value with respect to point $\mathbf{p}$:
\begin{equation} \label{equ:1}
n(\mathbf{p}) = \frac{\partial s(\mathbf{p})}{\partial {\mathbf{p}}},
\end{equation}
where $n(\mathbf{p})$ is the surface normal vector at the point~$\mathbf{p}$.

\section{Method}

We extract geometric priors for each of the given RGB images using OmniData~\cite{eftekhar2021omnidata}, and obtain sparse 3D points and camera poses using COLMAP~\cite{schonberger2016structure}. To integrate the information provided by these priors, we construct multi-level feature volume grids and MLP decoders for SDF and color, respectively. We then use trilinear interpolation to sample the features of query points along camera rays, and render the results with the MLP decoders. As all of these processes are differentiable, we can optimize the feature grids and decoders by supervising them with our loss functions. An overview of our method can be seen in Figure \ref{Fig:architecture}.
\begin{figure}
    \centering
    \includegraphics[width=0.90\linewidth]{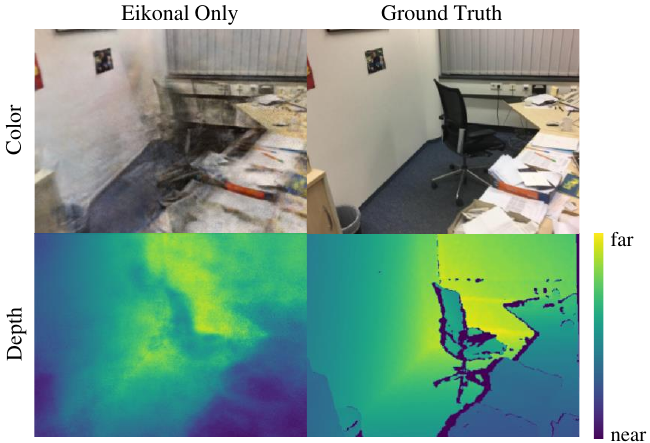}
    \caption{
        Qualitative results in a real-world scene. It shows that the Eikonal loss has difficulties in reconstructing surface geometry from a few training images which results in over-smooth depth and color rendering qualities. Top : images, bottom : depth maps.
    }
    \label{Fig:fail_eikonal}
\end{figure}

\subsection{Conventional Surface Regularization Loss}
\label{sec: geometric_regularization_loss}

For general implicit neural surface reconstruction tasks with signed distance functions (SDF)~\cite{wang2022go, azinovic2022neural, wang2021neus, yariv2021volume, zhu2023nicer}, Eikonal loss~\cite{gropp2020implicit} is commonly used for the entire neural implicit field to regularize geometry.
By utilizing the property of SDF, the Eikonal loss enforces the gradient of SDF to be a constant value of 1, as in,
\begin{equation}
L_{\text{Eik}} = \sum_{\mathbf{p} \in R} \Big(\lVert n(\mathbf{p}) \lVert_2 - 1 \Big)^2,
\end{equation}
where $R$ is a set of rays.

However, Eikonal loss is effective when the nerual implicit representation can build detailed surface with abundant information from various viewpoints.
When only sparse input views are given, it is difficult to retrieve reliable 3D geometry because there are many possible convergence points due to ambiguity. 
As was mentioned in the introduction section, MonoSDF~\cite{yu2022monosdf} showed that using Eikonal loss in few-shot NeRF training encounters unstable optimization during reconstructing and synthesizing novel views.
It is also shown in Figure~\ref{Fig:fail_eikonal} that when utilizing voxel grid optimization for fast training in few-shot setup, using only Eikonal loss brings low-fidelity results of color images and depth maps.

Therefore, in order to use surface regularization technique in fast few-shot NeRF training, we should consider a way to give reliable surface supervision, \ie, to build \textit{surface-friendly} environment.
The following question would be, how can we build \textit{surface-friendly} environment from sparse input images?

\subsection{Annealing Signed Distance Function Loss}

\begin{figure}
    \centering
    \includegraphics[width=0.95\linewidth]{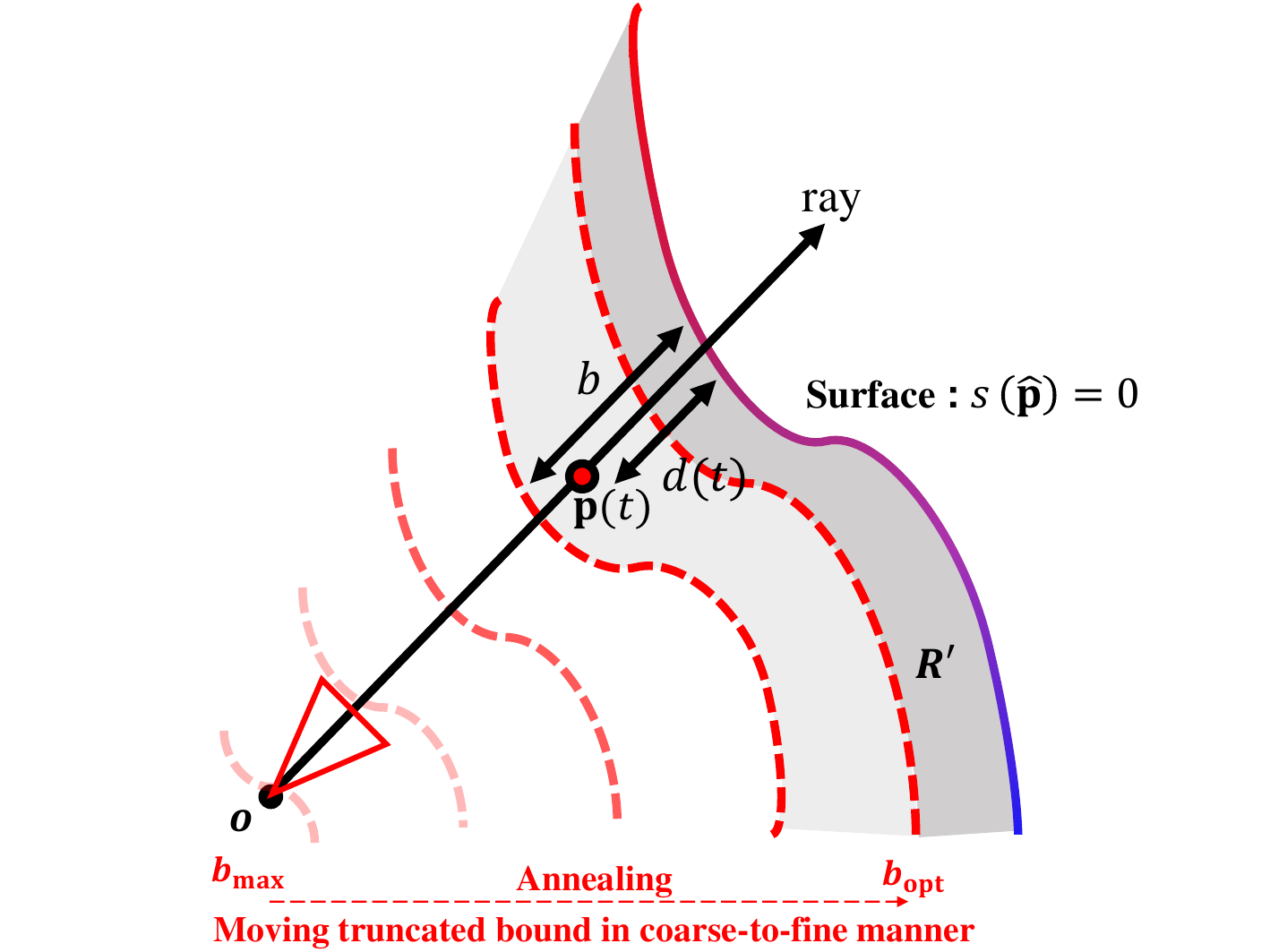}
    \caption{
        The visualization for Annealing SDF loss. The truncated bounds~$b$ decrease from $b_\text{max}$, camera origin $o$, to $b_\text{opt}$ by following iteration process to capture detailed geometry by reducing smoothing area. $L_\text{GS}$ is applied to $\textbf{p}(t)$ within \textcolor{grey}{grey} region $R'$.
    }
    \label{Fig:gsdf_explain}
\end{figure}

To remedy this problem, we choose coarse-to-fine strategy in surface regularization.
We propose Annealing Signed Distance Function (ASDF) loss to do so.
Our ASDF loss aims to enforce adaptive geometric smoothing in regard to the training steps.
In the earlier stage of training, ASDF loss guides the network to learn overall structure, while in the latter stage, the weight of smoothing loss is decreased so that the network can focus on refining details. We visualize the overview for ASDF loss in Figure~\ref{Fig:gsdf_explain}.

Our ASDF loss consists of two terms, which are geometric smoothing loss $L_\text{GS}$ and weighted Eikonal loss $L_\text{Eik}$.
Geometric smoothing loss, which is the core component of our ASDF loss, is defined as, 
\begin{equation}
L_{\text{GS}} =
\begin{cases}
\sum_{\mathbf{p}(t) \in R'}|{s(\mathbf{p}(t)) - d(t)}| & b > d(t) \\
0 & b \leq d(t)
\end{cases}.
\end{equation}
Here, $d(t)$ is $\hat{D}_r - t$, which refers to the distance between query point and the rendered depth value from a given camera ray.
Additionally, $b$ is the truncated bound and $R'$ is region between surface and moving truncated bound.
Therefore, the geometric smoothing loss enforces SDF value to be same as the distance between query point and the intersection of the camera ray and rendered surface.
By doing so, the resulted surface from estimated SDF should be smooth in truncated region.

To enable coarse-to-fine strategy, we adjust the bound $b$ during the optimization process.
$b$ decreases from $b_\text{max}$ to $b_\text{opt}$ as the training iteration increases, as in,
\begin{equation}
b =
\begin{cases}
\frac{i_{\text{opt}} - i}{i_{\text{opt}}} (b_{\text{max}} - b_{\text{opt}}) + b_{\text{opt}}& i \leq i_{\text{opt}} \\
b_{\text{opt}} & i > i_{\text{opt}}
\end{cases}.
\end{equation}

These equations make geometric smoothing in wider range of query points in the early training stage, and reduce this range gradually, so that the network first optimizes to coarse surface, and then progressively recover detailed geometry.

The second term, weighted Eikonal loss, is defined as was introduced in~\cite{wang2022go},
\begin{equation}
L_\text{wEik} = \frac{1}{\sum_{\mathbf{p} \in R}|s(\mathbf{p})|}\sum_{\mathbf{p} \in R} \Big( (\lVert n(\mathbf{p}) \lVert_2 - 1)^2 \cdot |s(\mathbf{p})| \Big).
\end{equation}

Thus, our annealing signed distance function (ASDF) is represented as $L_{\text{ASDF}} = L_\text{GS} + \lambda_\text{wEik} L_\text{wEik}$.

\subsection{Architecture}
\label{sec: architecture}

Our architecture consists of multi-level feature grids $V_{s}^i$ for SDF and $V_{c}^i$ for color, as well as MLP decoders $f_{s}$ and $f_{c}$ for SDF and color, respectively. Features are extracted from each feature volume grid by interpolating the nearest 8 points to normalized sampled query points. These extracted features are concatenated and fed into each decoder $f_{s}$ and $f_{c}$ with positional encoding~\cite{tancik2020fourier} to obtain the SDF value $s$ and color value $c$ at the query point $p$, as in,
\begin{equation}
s(\mathbf{p}) = f_s\big(\text{interp}(\mathbf{p}, V_s^i), \text{PE}(\mathbf{p})\big),
\end{equation}
\begin{equation}
c(\mathbf{p}) = f_c\big(\text{interp}(\mathbf{p}, V_c^i), \text{PE}(\mathbf{p})\big),
\end{equation}
where $\text{interp}(\cdot)$ is trilinear interpolation and $\text{PE}(\cdot)$ indicates positional encoding~\cite{mildenhall2021nerf}.

To optimize our network, we employed three rendering losses for each rendered value, as well as aforementioned ASDF loss.
The rendering losses for rendered color and rendered normal are based on the Euclidean distance between the ground truth and predicted values:
\begin{equation}
L_{C} = \sum_{r \in R} \lVert  C_r - \hat{C}_r \lVert_2,~~~L_{N} = \sum_{r \in R} \lVert  N_r - \hat{N}_r \lVert_2.
\end{equation}
$C_r$ refers to ground truth color and $N_r$ and $\hat{N}_r$ are surface normal predicted by OmniData and rendered surface normal, respectively.

For dense depth prediction, \cite{yu2022monosdf} use ground truth depth information to scale initial off-the-shelf depth estimation results.
In contrast, we do not utilize ground truth, therefore applied a least squares optimization to directly scale it to the projected sparse points from SfM, using a scale factor $w$ and a shift factor $b$.
To supervise the scaling, we used rendered depth in the following loss function:
\begin{equation}
L_{D} =
\begin{cases}
\sum_{r \in R} \lVert (wD_r + b) - \hat{D_r}\lVert_2 & N_{s} > 0 \\
0 & \text{otherwise}
\end{cases},
\label{equ:scaling}
\end{equation}
where $D_r$ is a predicted depth by OmniData. $N_{s}$ is the number of projected sparse points to the camera. Note that since some images may not contain any projected sparse points, $w$ and $b$ cannot be solved and are therefore not supervised with depth.
 The total loss for optimization for our architecture is given by:
\begin{equation}
L_{total} = L_C + \lambda_N L_N + \lambda_D L_D + \lambda_{\text{ASDF}} L_{\text{ASDF}}.
\end{equation}
\section{Experiments}

\begin{figure*}[ht]
    \centering
    \includegraphics[width=\textwidth]{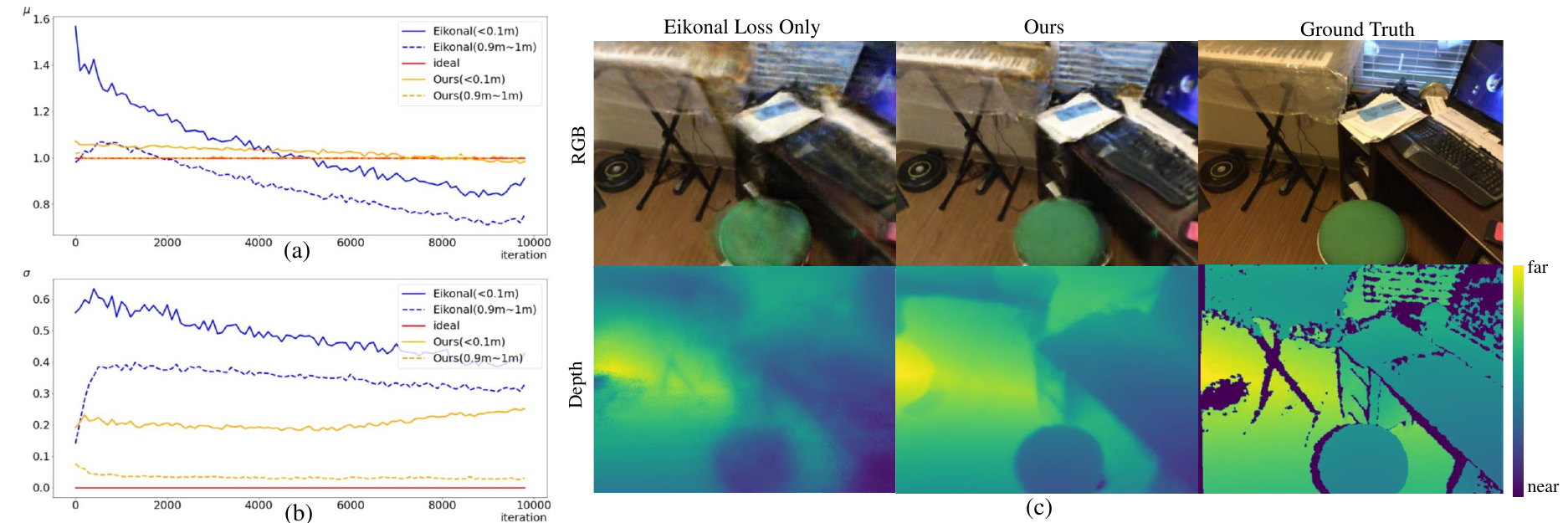}
    \caption{
        Analysis on geometric regularization loss. For both the Eikonal loss~(\textcolor{blue}{blue}) and our loss function~(\textcolor{orange}{orange}), we plotted the mean (a) and standard deviation (b) for the gradient at the nearby surface ($<$ 10cm) and far from the surface (90cm $\sim$ 1m) against iterations with ideal value~(\textcolor{red}{red}). In addition, we provided visualizations of the rendered RGB and depth maps for both cases in (c). The left images show the rendered results supervised only by the Eikonal loss for the SDF loss, while the middle images are supervised by our annealing SDF loss. The last column shows the ground truth images. In (c), top row : color images, bottom row : depth maps.
    }
    \label{Fig:gtsdf0}
\end{figure*}

\subsection{Implementation Details}
We implemented our system using a single TITAN RTX 24GB and adjusted the voxel sizes for each scene to fit the GPU memory and improve performance. We utilized $i=4$ for both SDF feature grids and color feature grids in multi-resolution feature grids.
Our decoders, $f_c$, and $f_s$, were implemented using MLPs with a single layer containing 128 channels.
The learning rates for the decoder and feature grids were set to 0.001 and 0.01, respectively.
For the initial value of standard deviation for the surface reconstruction parameter, we used 0.3 and a learning rate of 0.001, similar to \cite{wang2021neus}. We used the Adam optimizer \cite{kingma2014adam} to optimize our system.
To optimize our network, we set the following values for [$\lambda_N$, $\lambda_D$, $\lambda_\text{ASDF}$, $\lambda_\text{wEik}$, $i_{opt}$]: [0.1, 0.12, 0.2, 0.75, 150]. The truncated bounds [$b_\text{max}$, $b_\text{opt}$] were dependent on the scene scale.
Since PyTorch does not support differentiable grid sampling, we employ a customized grid sampler \cite{wang2022go}. 

\subsection{Dataset and Few-Shot Training Scheme}
\label{sec:Dataset}

The ScanNet dataset~\cite{dai2017scannet} consists of RGB-D Sequences in indoor environments and contains over 2.5 million images from more than 1000 real-world indoor scans. We selected five scenes from the ScanNet dataset and used 18 to 20 images from each pair of scenes that share common surfaces for training, as well as 8 images for validation and testing, following the protocol used in \cite{roessle2022dense}. We did not use dense measurements from the sensor, but rather relied on the results of SfM and deep dense priors to match scale.
The NeRF-Real dataset~\cite{mildenhall2019local, mildenhall2021nerf} includes eight forward-facing scenes. For each scene, we accessed 2 views, 5 views and 10 views to train our network. We followed the official protocol to evaluate the performance of our method on the NeRF-Real dataset~\cite{deng2022depth}.

\subsection{Impact of utilizing ASDF Loss}

\noindent\textbf{Stable optimization.}
In Figure~\ref{Fig:gtsdf0}-(a,b), while the mean and standard deviation for the Eikonal loss shows unstable, ours shows stable optimization process. By supporting geometric smoothing effect from ASDF loss, our approach dramatically decrease the instability of initial optimization process with the Eikonal loss. Also, we observed that unstable training affects the qualitative results. As shown in Figure~\ref{Fig:gtsdf0}-(c), using only Eikonal loss results in unreliable color and depth rendering. 
Thanks to our geometric smoothing term, our approach help network be able to make clear color image and geometry.

\begin{figure}
    \centering
    \includegraphics[width=0.95\linewidth]{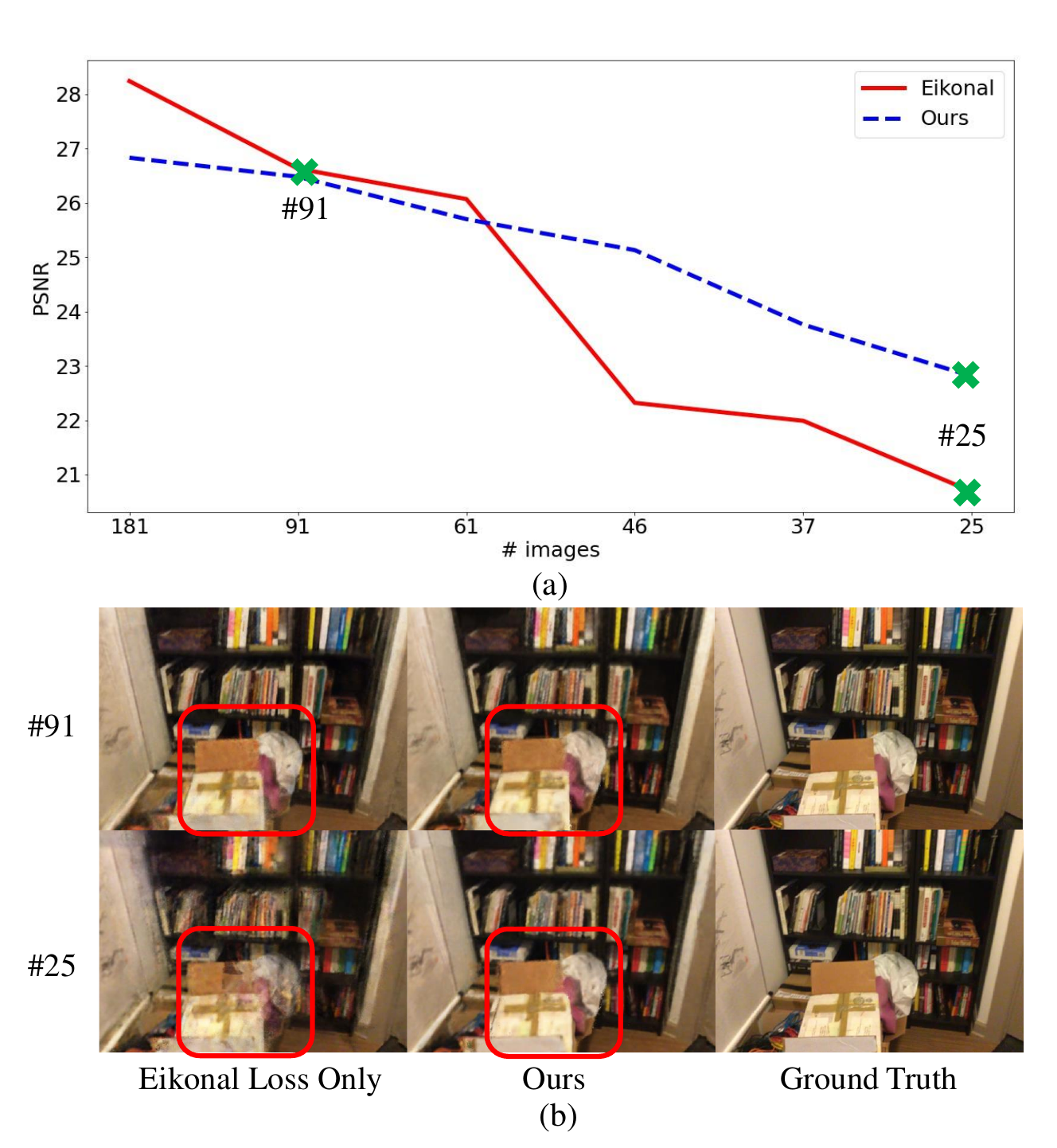}
    \caption{
        Analysis on two geometric regularizations in terms of number of training images. (a) We plot the relationship between the number of training images and the rendered image quality using PSNR ($\uparrow$) metric. (b) sampled rendering results with ground truth images. Note that $\#$91 (top row) indicate the results using 91 images and similarly for $\#$25 (bottom row).
    }
    \label{Fig:ablation_n_img}
\end{figure}

\noindent\textbf{Few-shot learning.}
We examined the relationship between the number of training images and the performance for novel view synthesis using our training scheme.

We used PSNR to measure performance.
As shown in Figure \ref{Fig:ablation_n_img}-(a), due to the smoothing effect of our proposed method, our performance was slightly lower than the naive approach when a lot of images are used for training. However, we observed a significant performance drop at certain number of images with the Eikonal loss, whereas our approach exhibited a stable performance decrease. In the qualitative results presented in Figure \ref{Fig:ablation_n_img}-(b), the upper row depicts novel view synthesis with 91 training images, and the lower row shows synthesis with 25 training images. Despite the reduced number of images, our approach showed no significant difference in visualization. In contrast, the naive approach failed to synthesize reliable novel view.

\subsection{Comparison with State-of-the-Arts}
\noindent\textbf{Baselines and metrics.}
We compared our method with baseline approaches for novel view synthesis, namely NeRF~\cite{mildenhall2021nerf} and state-of-the-art few-shot NeRF with SfM: DS-NeRF~\cite{deng2022depth}, DDP-NeRF~\cite{roessle2022dense} and SCADE~\cite{uy2023scade}. DS-NeRF utilized naive results of SfM and used the error as weight for supervision. DDP-NeRF utilized SfM to predict dense depth maps and uncertainty. SCADE exploited ambiguity-aware depth estimates to address inherent ambiguities of monocular depth estimation in-the-wild indoor scenes. 

To evaluate the quality of the rendered images, we employed three color evaluation metrics: PSNR, SSIM, and LPIPS, as well as a single geometric evaluation metric for rendered depth, RMSE error, as following~\cite{roessle2022dense}. PSNR measures the color difference between two images, indicating the overall quality of the images. In contrast, SSIM is used to evaluate structural similarity between two images. Both evaluation metrics lack perception loss, which was addressed by \cite{zhang2018unreasonable} through the introduction of LPIPS, a metric that passes images to a pre-trained network and measures the difference between feature maps. 

\begin{figure*}[]
    \centering
    \includegraphics[width=\linewidth]{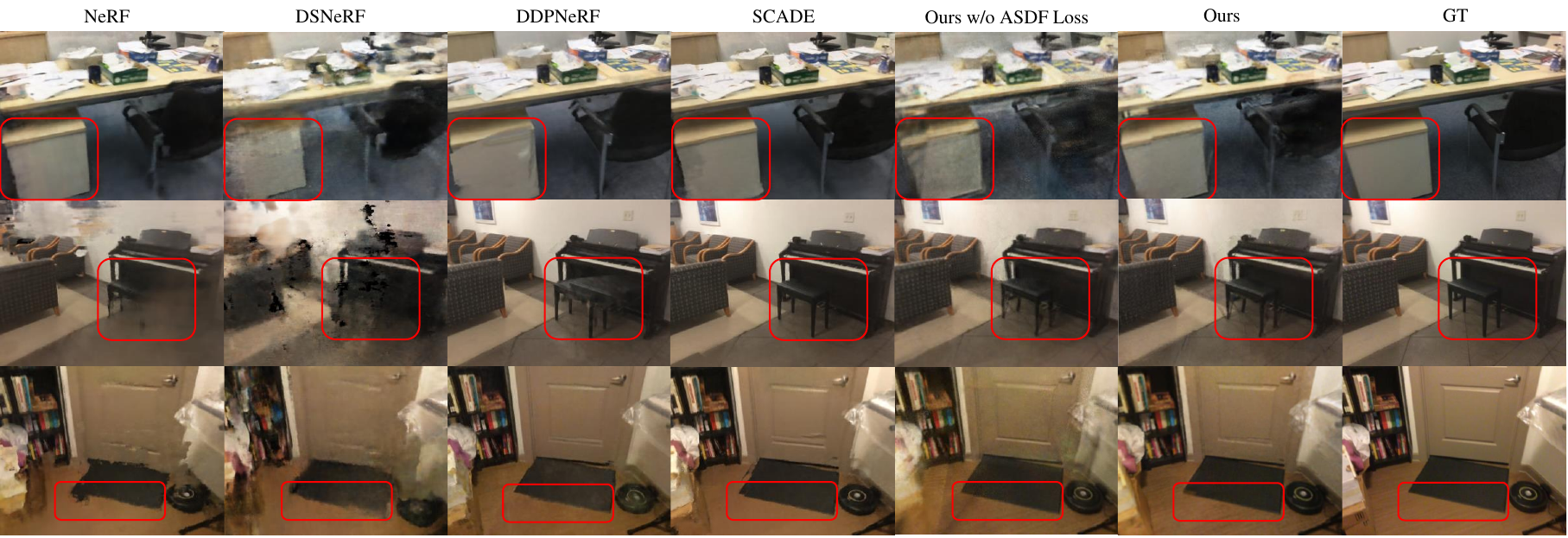}
    \caption{Qualitative comparison in the ScanNet dataset~\cite{dai2017scannet}. We utilized official PyTorch Implementations for DSNeRF~\cite{deng2022depth}, DDPNeRF~\cite{roessle2022dense} and SCADE~\cite{uy2023scade}. In contrast to alternative methods, our approach exhibits greater robustness on planar surfaces such as drawers and floors.}
    \label{Fig:scannet_qual0}
\end{figure*}

\begin{table}[!t]{
\resizebox{\linewidth}{!}{%
\begin{tabular}{l|c|cccc}

\Xhline{2\arrayrulewidth} 

 & Time $\downarrow$ & PSNR $\uparrow$ & SSIM $\uparrow$ & LPIPS $\downarrow$ & RMSE $\downarrow$ \\
\hline
NeRF & 23h & 19.03 & 0.670 & 0.398 & 1.163\\
DS-NeRF & 22h & 20.85 & 0.713 & 0.344 & 0.447\\
DDPNeRF & 15h & 20.96 & \textbf{0.737} & 0.294 & 0.236\\
SCADE & 13h & 21.54 & 0.732 & \textbf{0.292} & 0.253 \\
Ours w/o ASDF & \textbf{0.5h} & 20.52 & 0.633 & 0.408 & 0.307 \\
Ours & \textbf{0.5h} & \textbf{21.57} & 0.720 & \textbf{0.292} & \textbf{0.228}\\ 

\Xhline{2\arrayrulewidth} 

\end{tabular}
}}
\caption{Quantitative comparison in the ScanNet dataset~\cite{dai2017scannet}. Ours demonstrates plausible performance with a significantly improved optimization time. To compute the RMSE error of SCADE~\cite{uy2023scade}, we employed the official PyTorch implementation.}
\label{Table:eval_scannet}
\end{table}

\noindent\textbf{ScanNet dataset.}
We evaluated our method on ScanNet (Figure \ref{Fig:scannet_qual0}) with NeRF~\cite{mildenhall2021nerf}, DSNeRF~\cite{deng2022depth}, DDPNeRF~\cite{roessle2022dense}, SCADE~\cite{uy2023scade} and Ours without ASDF loss, which is utilizing Eikonal loss.
We selected three scenes and visualized the rendered color images. The visualization results illustrated in Figure~\ref{Fig:scannet_qual0} show that our approach is more robust in homogeneous regions, such as a chair or floor.

For quantitative evaluation, we used the same scene as \cite{roessle2022dense}. According to the results presented in Table \ref{Table:eval_scannet}, our approach shows promising performance in most metrics compared to the previous state-of-the-arts.
Note that while maintaining comparable results, ours accomplished a speed boost of 30 times compared to DDPNeRF~\cite{roessle2022dense}, and up to 45 times compared to NeRF~\cite{mildenhall2021nerf}.

\begin{figure*}
    \centering
    \includegraphics[width=\linewidth]{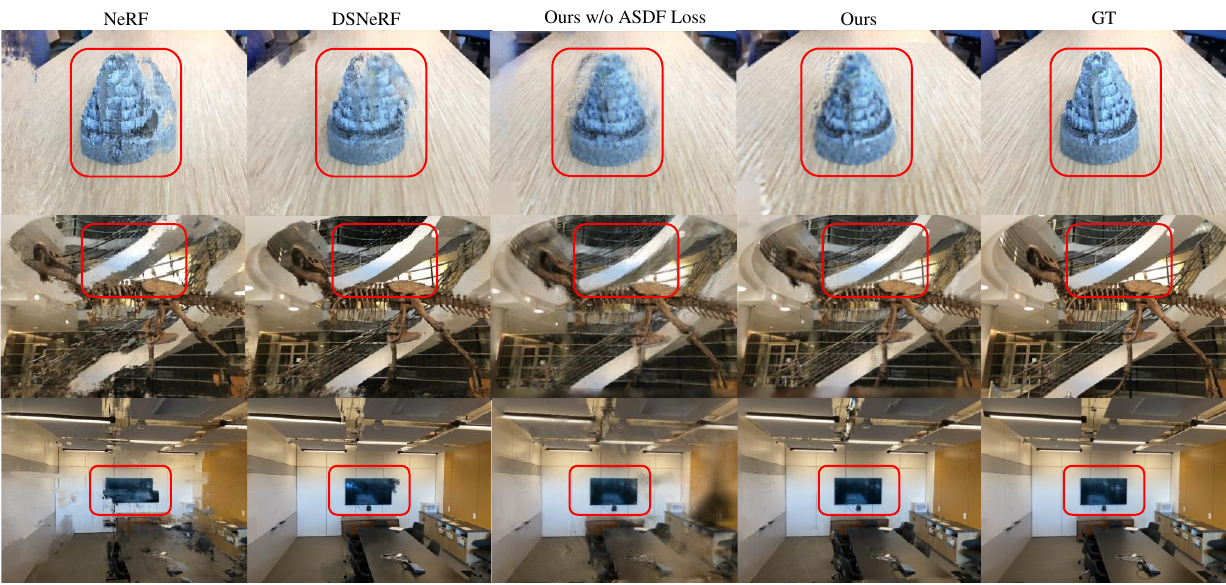}
    \caption{Qualitative comparison in NeRF-Real dataset~\cite{mildenhall2019local, mildenhall2021nerf}. Ours demonstrates plausible results, in contrast with other methods that exhibit blurry effects or unstable shapes, such as TVs and handrails.}
    \label{Fig:nerf_real_qual0}
\end{figure*}

\noindent\textbf{NeRF-Real dataset.}
To show our ability in limited viewing direction scenes, we evaluated our method on the NeRF-Real dataset with NeRF~\cite{mildenhall2021nerf} and DSNeRF~\cite{deng2022depth} (Figure \ref{Fig:nerf_real_qual0}). We used every 8-th image for testing and trained with 2 views following \cite{deng2022depth}. Ours shows plausible visualization results when compared to another approaches.

For quantitative result in NeRF-Real dataset, we evaluate our approach with the average value for PSNR, SSIM and LPIPS of 2 views, 5 views and 10 views data split, following that of DSNeRF~\cite{deng2022depth}. Results are shown in Table~\ref{Table:eval_llff}.

Similar to the results from ScanNet dataset, our method outperforms vanilla NeRF in the most of the metrics in every scenes.
Compared to DSNeRF, our approach shows comparable results on PSNR and SSIM metrics.
Please note that in LPIPS metric, ours dramatically outperforms both NeRF and DSNeRF.
From this observation, we believe that our proposed fast few-shot NeRF training scheme enables reliable radiance optimization that is sufficient to feature retrieval, with up to x45 training time.

\begin{figure}
    \centering
    \includegraphics[width=\linewidth]{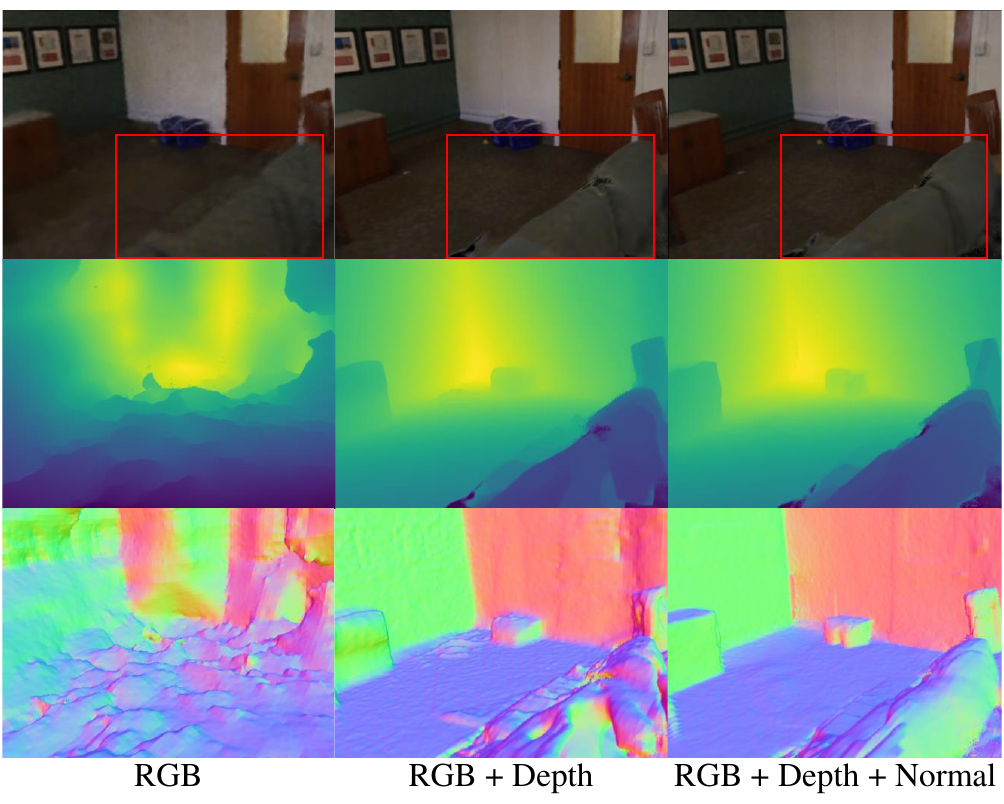}
    \caption{
        Ablation study for geometry priors with our geometric smoothing loss. Top: rendered color images, mid: rendered depth maps, bottom: rendered surface normal maps.
    }
    \label{Fig:scannet_ablation}
\end{figure}

\begin{table}[!t]{
\resizebox{\linewidth}{!}{%
\begin{tabular}{ccc|cccc}

\Xhline{2\arrayrulewidth} 

\multicolumn{3}{c|}{Preserve (\checkmark)} & \multicolumn{4}{c}{Evaluation} \\

\hline
RGB & Depth & Normal & PSNR $\uparrow$ & SSIM $\uparrow$ & LPIPS $\downarrow$ & RMSE $\downarrow$ \\

\hline
\checkmark & & & 16.88 & 0.501 & 0.477 & 1.693 \\
\checkmark & \checkmark &  & 21.07 & 0.704 & 0.307 & 0.239\\
\checkmark & \checkmark & \checkmark & \textbf{21.17} & \textbf{0.708} & \textbf{0.303} & \textbf{0.228}\\

\Xhline{2\arrayrulewidth} 

\end{tabular}
}}
\caption{Ablation study for dense predictions with our ASDF loss.}
\label{Table:ablation_denseprediction}
\vspace{-3mm}
\end{table}

\begin{table*}[!t]{
\centering
\resizebox{\textwidth}{!}{%
\begin{tabular}{l|ccc|ccc|ccc|ccc}

\Xhline{2\arrayrulewidth} 

& &    NeRF (Time: 23h)    & & &    DSNeRF (Time: 22h)    & & & \hspace{-5mm} Ours w/o ASDF (Time: \textbf{0.5h})  \hspace{-5mm} & & & \hspace{5mm}  Ours (Time: \textbf{0.5h}) \hspace{5mm}\\

& PSNR $\uparrow$ & SSIM $\uparrow$ & LPIPS $\downarrow$ & PSNR $\uparrow$ & SSIM $\uparrow$ & LPIPS $\downarrow$ & PSNR $\uparrow$ & SSIM $\uparrow$ & LPIPS $\downarrow$ & PSNR $\uparrow$ & SSIM $\uparrow$ & LPIPS $\downarrow$ \\

\hline

fern & 20.00 & 0.58 & 0.41 & \textbf{20.60} & \textbf{0.60} & 0.43 & 20.19 & 0.51 & 0.43 & 20.26 & 0.56 & \textbf{0.34}\\
flower & 17.79 & 0.49 & 0.43 & \textbf{21.79} & \textbf{0.62} & 0.35 & 19.87 & 0.54 & 0.31 & 19.27 & 0.51 & \textbf{0.29}\\
fortress & 19.13 & 0.55 & 0.53 & 23.85& 0.66 & 0.35 & 22.93 & 0.66 & 0.21 &\textbf{24.08} & \textbf{0.67} & \textbf{0.18} \\
horns & 17.57 & 0.51 & 0.50 & \textbf{18.86}& \textbf{0.53} & 0.54& 18.13 & 0.47 & 0.44& 17.85 & 0.50 & \textbf{0.39} \\
leaves & 13.73 & 0.26 & 0.52 & 15.93 & 0.35 & 0.5& \textbf{16.90} & \textbf{0.48} & 0.33 & 16.54 & 0.45 & \textbf{0.32} \\
orchids & 14.79 & 0.34 & 0.45 & 16.15 & 0.39 & 0.44 & \textbf{16.16} & 
0.37 & 0.39 & 15.84 & \textbf{0.42} & \textbf{0.27} \\
room & 22.23 & 0.78 & 0.32 & \textbf{24.46} & \textbf{0.86} & \textbf{0.19} & 22.22 & 
0.74 & 0.28 &23.22 & 0.81 & 0.20 \\
trex & 18.34 & \textbf{0.63} & 0.34 & 18.82 & \textbf{0.63}  & 0.39 & 19.04 & 0.56 & 0.39 & \textbf{19.07} & 0.59 & \textbf{0.30} \\

\hline

Avg. & 17.95 & 0.52 & 0.42 & \textbf{20.06} & \textbf{0.58} & 0.40 & 19.43 & 0.54 & 0.35 & 19.51 & 0.56 & \textbf{0.29}\\

\Xhline{2\arrayrulewidth} 

\end{tabular}
}}
\vspace{1mm}
\caption{Quantitative comparison in NeRF-Real Dataset~\cite{mildenhall2019local, mildenhall2021nerf}. The results are the average of 2 views, 5 views, and 10 views following DSNeRF~\cite{deng2022depth}.}
\label{Table:eval_llff}
\vspace{-2mm}
\end{table*}

\subsection{Ablation Study}


\noindent\textbf{Utilization of deep dense priors.}
In order to evaluate the effectiveness of deep dense priors which are utilized in our approach, we ablate a few variants that employ different types and numbers of 3D modalities with our ASDF loss.

The task of producing accurate dense predictions with RGB images only is challenging, as demonstrated in Figure \ref{Fig:scannet_ablation}. However, with the incorporation of deep dense priors with supervision, we observed a significant improvement in all metrics, as indicated in Table~\ref{Table:ablation_denseprediction}.

\section{Conclusion}

We have presented a fast few-shot NeRF with deep dense priors and structure from motion. Given the difficulty of optimizing geometric information from a few views in complex scenes, we propose a new surface regularization loss, \textit{Annealing Signed Distance Function loss}, that enforces geometric smoothing and improves the performance for synthesizing novel view. As a result, we have successfully connected deep dense priors, multi-view consistency and multi-resolution voxel grids for novel view synthesis with sparse input views.

Our method can be further enhanced by adapting recent approaches such as \cite{muller2022instant, fridovich2022plenoxels} to improve the optimization speed of the NeRF. Also, uncertainty handling for geometric priors may boost the performance by reducing the error of off-the-shelf network.

For the limitation of this method, we think that our Annealing Signed Distance Function requires hyper-parameters that are dependent on the scene geometry or SfM results, \eg, accuracy of camera poses. We think that solving this issue in an adaptive manner without heuristic tuning can be a future direction.

\noindent \textbf{Acknowledgements} This work was supported by the National Research Foundation of Korea (NRF) grant funded by the Korea government (MSIT) (NRF2022R1A2B5B03002636).

{
    \small
    \bibliographystyle{ieeenat_fullname}
    \bibliography{main}
}

\end{document}